\pdfoutput=1

\documentclass[11pt]{article}

\usepackage[final]{acl}

\usepackage{times}
\usepackage{latexsym}

\usepackage[english]{babel}
\usepackage{graphicx}
\usepackage{amsmath}
\usepackage{booktabs}
\usepackage{graphicx}
\usepackage{array}
\usepackage{multirow}
\usepackage{float} 
\usepackage{subfigure} 
\usepackage{enumitem}

\usepackage[T1]{fontenc}

\usepackage[utf8]{inputenc}

\usepackage{microtype}

\usepackage{inconsolata}

\title{Encoding Hierarchical Schema via Concept Flow for\\
	Multifaceted Ideology Detection}

\author{Songtao Liu\textsuperscript{\textnormal{1}},
	Bang Wang\textsuperscript{\textnormal{1,}}\footnotemark[1],
	Wei Xiang\textsuperscript{\textnormal{1}},
	Han Xu\textsuperscript{\textnormal{2}} and
	Minghua Xu\textsuperscript{\textnormal{2,}}\Thanks{\ Corresponding author: B. Wang and M. Xu}
	\\
	\textsuperscript{1}School of Electronic Information and Communications, \\ Huazhong University of Science and Technology, Wuhan, China \\
	\textsuperscript{2}School of Journalism and Information Communication, \\ Huazhong University of Science and Technology, Wuhan, China \\
	\texttt{\{liusongtao, wangbang, xiangwei, xuh, xuminghua\}@hust.edu.cn}}

\begin{document}
\maketitle
\begin{abstract}
Multifaceted ideology detection (MID) aims to detect the ideological leanings of texts towards multiple facets. Previous studies on ideology detection mainly focus on one generic facet and ignore label semantics and explanatory descriptions of ideologies, which are a kind of instructive information and reveal the specific concepts of ideologies. In this paper, we develop a novel concept semantics-enhanced framework for the MID task. Specifically, we propose a bidirectional iterative concept flow (BICo) method to encode multifaceted ideologies. BICo enables the concepts to flow across levels of the schema tree and enriches concept representations with multi-granularity semantics. Furthermore, we explore concept attentive matching and concept-guided contrastive learning strategies to guide the model to capture ideology features with the learned concept semantics. Extensive experiments on the benchmark dataset show that our approach achieves state-of-the-art performance in MID, including in the cross-topic scenario.\footnote{\ The source code is available at \url{https://github.com/LST1836/BICo}}
\end{abstract}

\section{Introduction}

\begin{figure}[t] 
	\centering 
	\includegraphics[width=\columnwidth]{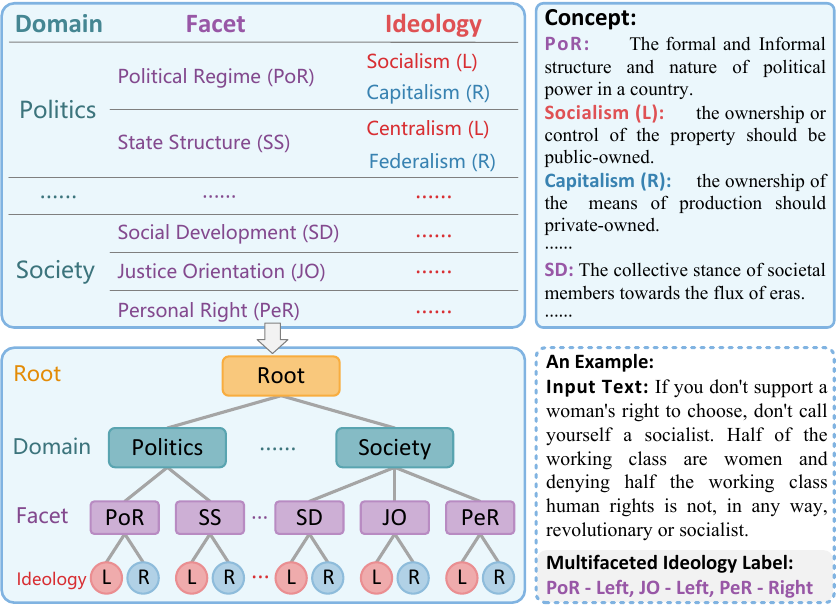} 
	\caption{Upper: the multifaceted ideology schema and concepts of facets and ideologies \citep{liu-etal-2023-ideology}. Lower left: the tree-like hierarchical structure of the schema. Lower right: an example of MID. ``L'' denotes Left, ``R'' denotes Right.} 
	\label{fig: intro} 
\end{figure}

Multifaceted ideology detection (MID) aims to identify the ideological leanings (e.g., Left, Center, Right, etc.) expressed in texts towards multiple facets, as shown in Figure~\ref{fig: intro}. It is crucial for understanding public opinion and detecting potential extremism \citep{kannangara2018mining, grover2019detecting, demszky-etal-2019-analyzing}, which is helpful for governments and cybersecurity organizations \citep{stefanov-etal-2020-predicting, aldera2021online}. It can also facilitate downstream research and applications in social sciences \citep{kabir2022deep}.

In most of related work, researches generally focus on modeling the text content with diversified cues, such as sentiment polarities \citep{bhatia-p-2018-topic, kabir2022deep}, named entities \citep{liu-etal-2022-politics}, and discourse structure \citep{devatine-etal-2023-integrated, hong-etal-2023-disentangling}, or jointly learning with other related tasks \citep{baly-etal-2019-multi}. There are also approaches that incorporate information sources beyond text to facilitate ideology mining. Hyperlink structure \citep{kulkarni-etal-2018-multi}, social networks \citep{stefanov-etal-2020-predicting, li-goldwasser-2021-using}, external knowledge from knowledge graphs \citep{zhang-etal-2022-kcd} as well as information from other modalities \citep{qiu-etal-2022-late}, are introduced in the task of ideology detection.

Although achieving promising performance, those methods limit the ideology prediction to a generic facet. In other words, they only label a text as ideologically left- or right-leaning as a whole, regardless whether the text containing one or more different facets. Furthermore, they ignore a crucial clue, label semantics, that is, what exactly does an ideology mean? In this case, ideological categories are represented as one-hot vectors without any semantic information, and models can only rely on the training data distribution to analyze latent ideology features, which could be unfavorable for the generalization ability of models \citep{wang-etal-2021-concept, wen-hauptmann-2023-zero}.

So how can we effectively detect \textit{multifaceted} ideologies? And what exactly does an ideology mean? \citet{liu-etal-2023-ideology} propose the multifaceted ideology detection task for the first time and design a multifaceted ideology schema which contains 12 facets covering 5 domains in a  tree-like hierarchical structure (see Figure~\ref{fig: intro}, details in Appendix~\ref{sec: schema}). Each facet, as well as the ideological attributes under each facet, are defined using natural texts, which can be regarded as \textit{concepts}. These \textit{concepts} describe the meaning of facets and ideologies, thus making it natural to represent the label semantics. In addition, in the hierarchical schema, higher-level concepts (like Domain and Facet) have general semantics shared by their child concepts, while lower-level concepts (like Ideology and Facet) describe their parents from various views, which can be seen as the semantic divisions of higher-level concepts. This meaningful hierarchical structure can be utilized to enrich the concept semantics.

Based on the motivation above, to incorporate the concept semantics and leverage the hierarchical structure of the schema in MID, we propose a novel \textbf{B}idirectional \textbf{I}terative \textbf{C}oncept Fl\textbf{o}w (\textbf{BICo}) method to encode the hierarchical schema. Specifically, BICo allows concepts to flow in two directions on the schema tree, enabling them to perceive both high-level general semantics and low-level specific perspectives. On the one hand, inspired by the relation rotation in complex space \citep{sun2018rotate}, we design Concept Metapath Diffusion to perform message passing from root to leaf. On the other hand, in the direction of leaf to root, we propose Concept Hierarchy Aggregation to aggregate concept semantics in lower levels to the ones in higher levels based on the parent-child relation. Concept flow in the two directions is iterated multiple times and the final concept representations are enriched by multi-granularity semantics. For example, the Facet representations capture the meanings of different ideologies in the corresponding facet, while the Ideology representations also perceive information about the Facet and Domain they belong to. We match the text and Facet representations based on the attention mechanism to recognize text-related facets. Furthermore, we explore a Concept-Guided Contrastive Learning strategy to learn more distinguishable text representations under the guidance of Ideology concepts.

The main contributions of our work are summarized as follows:

(1) We propose a concept semantics-enhanced MID framework. To our best knowledge, this is the first work that incorporates label semantics and explanatory descriptions in the MID task.

(2) We propose a Bidirectional Iterative Concept Flow (BICo) method to encode the hierarchical schema. Concepts flow on the schema tree in two directions iteratively to capture multi-granularity concept semantics.

(3) We design Concept Attentive Matching and Concept-Guided Contrastive Learning strategies to enable the model to extract ideology features with the help of concept semantics.

(4) Extensive experiments on the MITweet benchmark demonstrate the effectiveness of our approach, including in the cross-topic scenario.

\begin{figure*}[t] 
	\centering 
	\includegraphics[width=\textwidth]{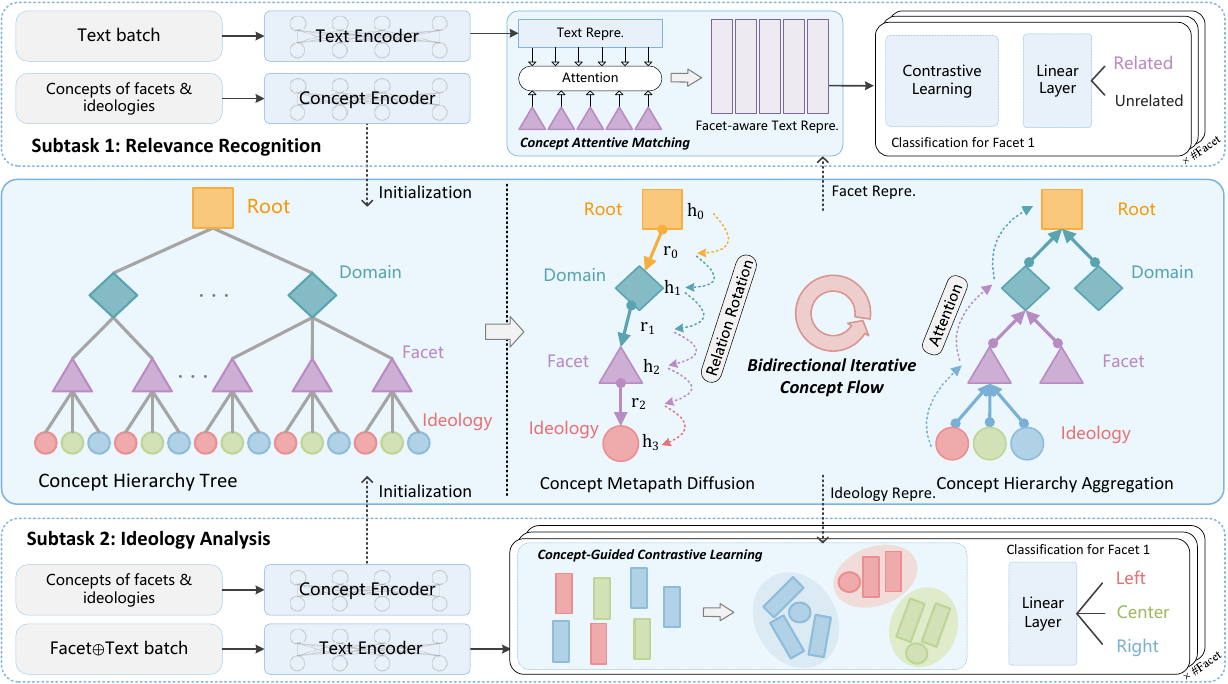} 
	\caption{Overview of our concept-enhanced multifaceted ideology detection framework. The blue box in the middle shows the proposed bidirectional iterative concept flow (BICo), which includes root-to-leaf concept metapath diffusion and leaf-to-root concept hierarchy aggregation. The concept representations are enriched gradually by bidirectional iteration, and are then used to enhance the two subtasks of MID through concept attentive matching and concept-guided contrastive learning.} 
	\label{fig: model} 
\end{figure*}

\section{Task Description}
Given an input text and a set of facets, Multifaceted Ideology Detection (MID) is divided into two sub-tasks: (1) \textbf{Relevance Recognition} aims to recognize the facets that the text is related to; (2) \textbf{Ideology Analysis} predicts which ideology the text holds towards the \textit{related} facets. Formally, a sample instance can be considered as a triple $\left(x, \{y_R^i\}_{i=1}^n, \{y_I^i\}_{i=1}^m\right)$, where $x$ is the input text, $y_R^i \in $ $\{$Related, Unrelated$\}$ represents the relevance label of $i$-th facet, $n$ is the number of given facets. For each facet that the text is related to, we have an ideology label $y_I^i \in$ $\{$Left, Center, Right$\}$, $m$ is the number of related facets.

\section{Approach}

In this section, we first introduce the proposed \textbf{B}idirectional \textbf{I}terative \textbf{C}oncept Fl\textbf{o}w (\textbf{BICo}) for encoding the hierarchical schema, and then discuss how we augment multifaceted ideology detection based on the learned concept encodings. Figure~\ref{fig: model} illustrates the overall structure of our model.

\subsection{Bidirectional Iterative Concept Flow}

\subsubsection{Concept Hierarchy Tree}

\citet{liu-etal-2023-ideology} define the first hierarchical schema of multifaceted ideology, which contains 12 facets covering 5 domains. We construct a concept hierarchy tree $T=(N, E)$ based on the schema, as shown in Figure~\ref{fig: model}. The node set $N$ contains four types of nodes, i.e., \texttt{Root}, \texttt{Domain}, \texttt{Facet} and \texttt{Ideology}. The edge set $E$ indicates the subordination relation between nodes. The \texttt{Ideology}, or leaf, nodes represent the three ideologies (Left, Center, Right) of each facet.

To initialize node embeddings in the concept hierarchy tree, we leverage the concepts of facets and ideologies in the schema. Specifically, we adopt a pre-trained language model as the concept encoder and feed the concepts of facets and ideologies in the schema into the encoder. We then extract the hidden state of [CLS] token as initial representations of \texttt{Facet} and \texttt{Ideology} nodes, i.e., $\textbf{h}_F$ and $\textbf{h}_I$. For \texttt{Root} and \texttt{Domain} nodes, we obtain their initial embeddings ($\textbf{h}_R$ and $\textbf{h}_D$) by average-pooling their child node embeddings.

\subsubsection{Concept Metapath Diffusion}

In the concept hierarchy tree, higher-level nodes (like \texttt{Root} and \texttt{Domain}) have general and abstract concepts, which are shared by their child nodes and could be beneficial for enriching the representations of lower-level nodes (like \texttt{Ideology} and \texttt{Facet}). In order to allow lower-level nodes to perceive higher-level abstract semantics, we adopt the relation rotation in complex space \citep{sun2018rotate}, which is effective for information transfer along edges in a sequential structure.

Specifically, we define concept metapath as a path from root to leaf $(\texttt{Root}-\texttt{Domain}-\texttt{Facet}-\texttt{Ideology})$. Given node representations in a metapath $(\mathbf{h}_R, \mathbf{h}_D, \mathbf{h}_F, \mathbf{h}_I)=(\mathbf{h}_0, \mathbf{h}_1, \mathbf{h}_2, \mathbf{h}_3)$, let $\mathbf{r}_i$ be the representation of edge between node $h_i$ and $h_{i+1}$, the concept metapath diffusion from root to leaf through relation rotation is formulated as:
\begin{align}
	\mathbf{o}_0 &= \mathbf{h}_0' = \mathbf{h}_0 \\
	\mathbf{h}_i' &= \mathbf{h}_i + \mathbf{h}_{i-1}' \odot \mathbf{r}_{i-1} \\
	\mathbf{o}_i &= \frac{\mathbf{h}_i'}{i+1}
\end{align}
where $\mathbf{h}_i$, $\mathbf{r}_{i}$ and $\mathbf{o}_{i}$ are all complex vectors, $\mathbf{o}_{i}$ is the updated embedding, $\odot$ is the element-wise complex product and performs vector rotation in complex space. Here we can easily interpret a real vector of dimension $d$ as a complex vector of dimension $d/2$ by treating the first half of the vector as the real part and the second half as the imaginary part. We perform concept diffusion on all metapaths in the tree and $\mathbf{r}_{i}$ is shared between each two consecutive levels of nodes. Note that in relation rotation, edges represent the rotation angles of vectors in complex space. Therefore, $\mathbf{r}_{i}$ is first randomly initialized in the range of $(-\pi, \pi)$, and then its real and imaginary parts are obtained by the Euler's formula.

\subsubsection{Concept Hierarchy Aggregation}

In contrast to metapath diffusion, concept hierarchy aggregation enables concept flow from leaf to root. In the concept hierarchy tree, child nodes describe their parent node from different views, and thus can be regarded as more fine-grained concepts. Through concept hierarchy aggregation, we aggregate concept semantics of child nodes to their parent node, so as to enrich the representations of higher-level nodes.

We utilize the graph attention network (GAT), which aggregates features through attention mechanism in a graph. Considering the characteristics of concept hierarchy tree, we modify it to explicitly model the hierarchical structure and quantitatively measure the compatibility between hierarchies in the tree. Specifically, we only establish aggregation between the parent node and its own child nodes, which is different from aggregating over all one-hop neighbors in GAT. Additionally, we use different attention parameters at different levels to distinguish the aggregation features of each hierarchy.

Formally, for a parent node $p$ with embedding $\mathbf{h}_{p}$, we compute an aggregation weight for each child node $i$ and then weighted sum all child nodes' embeddings:
\begin{align}
	e_{pi} &= \textrm{LeakyReLU}\left(\mathbf{A}_{l}\left(\mathbf{h}_p\parallel\mathbf{h}_i \right)\right) \\
	\alpha_{pi} &= \frac{\textrm{exp}\left(e_{pi}\right)}{\sum_{j \in \mathcal{C}_p \cup \{p\}} 		\textrm{exp} \left(e_{pj}\right)} \\
	\mathbf{h}_p' &= \sigma\left(\sum_{i \in \mathcal{C}_p \cup \{p\}} \alpha_{pi}\mathbf{h}_i\right)
\end{align}
where $\mathbf{A}_{l}$ is the learnable parameter for aggregation of nodes in level $l$, $\mathcal{C}_p$ is the child node set of $p$, $\mathbf{h}_p'$ is the updated representation for $p$.

\subsubsection{Bidirectional Iteration}

The root-to-leaf metapath diffusion and leaf-to-root hierarchy aggregation are iterated multiple times to update node encodings. Finally, The new generated concept representations can be fully aware of higher-level general semantics and constructed with concepts from different aspects. Next we will enhance the MID task with the enriched Facet and Ideology representations, $\mathbf{c}_F$ and $\mathbf{c}_I$.

\subsection{Concept-Enhanced MID}

\subsubsection{Text Encoder}

We select a pre-trained language model as the text encoder. In the subtask of \textit{Relevance Recognition}, the encoder processes input sequence and outputs a hidden representation for each token: $\mathbf{X}=\{\mathbf{x}_i\}_{i=1}^L$, where $L$ is the length of text. For \textit{Ideology Analysis}, we concatenate the text and its \textit{related} facet concept, and then feed the sequence into text encoder to acquire the hidden state of [CLS] as text representation $\mathbf{t}$.

\subsubsection{Concept Attentive Matching}

In \textit{Relevance Recognition} subtask, to enable the text to be aware of label semantics (i.e., Facet concepts) and measure the importance of each token in relevance feature extraction, we adopt the cross-attention mechanism \citep{vaswani2017attention} to match the Facet and input token representations:
\begin{align}
	\mathbf{t}^i=\textrm{softmax} \left(\frac{\mathbf{c}_F^{i} \mathbf{X}^T}{\sqrt{d}}\right)\mathbf{X}
\end{align}
where $d$ is the dimension of vectors in the equation, the superscript $i$ represents $i$-th facet, $\mathbf{c}_F^{i}$ is $i$-th facet representation and $\mathbf{t}^{i}$ is $i$-th facet-aware text representation.

\subsubsection{Concept-Guided Contrastive Learning}
\label{sec:concept-guided cl}

To inject label semantics (i.e., Ideology concepts) into \textit{Ideology Analysis} subtask, we further explore a Concept-Guided Contrastive Learning strategy (CGCL), which tries to make intra-ideology representations more compact in the feature space and inter-ideology ones more distinguishable with the ideology concepts as anchors. The motivation is that ideology concepts describe the general meaning of ideologies. In the embedding space, this property can be interpreted as clustering, where an ideology concept anchor is the semantic center of samples with that ideological category.

Specifically, given text representations $\mathcal{B}=\{\mathbf{t}_i\}_{i=1}^B$ in a batch ($B$ is the batch size), and three Ideology representations $\mathcal{A}=\{\mathbf{c}_{I,L}, \mathbf{c}_{I,C}, \mathbf{c}_{I,R}\}$ (corresponding to Left, Center and Right respectively) which will be used as \textit{concept anchors} in the vector space, the concept-guided contrastive loss is formulated as:
\begin{gather}
	\mathcal{L}_{CGCL} = \frac{1}{3} \sum_{i \in \{L,C,R\}} \mathcal{L}_i \\
	\mathcal{L}_i = \log \frac{ {\textstyle \sum_{j \in \{j|y_{I,j}=i\}}} \exp\left(f\left(\mathbf{c}_{I,i},\mathbf{t}_j\right) / \tau\right) }{\sum_{\mathbf{v} \in \mathcal{B} \cup \mathcal{A} \backslash \{c_{I,i}\} } \exp\left(f\left(\mathbf{c}_{I,i},\mathbf{v}\right) / \tau\right)}
\end{gather}
where $y_{I,j}$ is the ideology label of $\mathbf{t}_j$, $f$ is the cosine similarity function, $\tau$ is temperature parameter. Note that $\mathcal{L}_{CGCL}$ is computed for each facet, and we omit the facet superscript for clarity.

\subsubsection{Classification and Training}

Considering the varying ideology features among different facets, we set up a classification head with a softmax function for each facet in both subtasks:
\begin{align}
	y^i = \textrm{softmax}\left(\mathbf{W}^{i} \mathbf{t}^{i} + \mathbf{b}^{i}\right)
\end{align}
where the superscript $i$ represents $i$-th facet, $\mathbf{W}^{i}$ and $\mathbf{b}^{i}$ are trainable parameters.

Note that in \textit{Relevance Recognition}, we also incorporate contrastive learning (CL), which is similar to the concept-guided CL in Sec.~\ref{sec:concept-guided cl}, but the anchors here are text representations themselves:
\begin{align}
	\mathcal{L}_{CL}=-\frac{1}{B} \sum_{i=1}^{B} \log \frac{\sum_{j \in \mathcal{B}_i} \exp \left( f \left( \mathbf{t}_{i},\mathbf{t}_{j} \right) / \tau \right)}{\sum_{k \in \{k|i \neq k\}} \exp \left( f \left( \textbf{t}_{i},\textbf{t}_{k} \right) / \tau \right)}
\label{eq11}
\end{align}
where $\mathbf{t}_{i}$ is the facet-aware text representation, $\mathcal{B}_i=\{j|i \neq j,y_{R,i}=y_{R,j}\}$, $y_{R,i}$ is the relevance label of $\mathbf{t}_{i}$, $B$ is the batch size, $\tau$ is temperature parameter. Here $\mathcal{L}_{CL}$ is also computed for each facet, and we omit the facet superscript for clarity.

Finally, the training loss of both subtasks is the weighted sum of cross-entropy classification loss and contrastive learning loss across all facets:
\begin{gather}
	\mathcal{L} = \frac{1}{n} \sum_{i=1}^{n} \left( \mathcal{L}_{CE}^{i} + \lambda \mathcal{L}_{CL}^{i} \right)
\end{gather}
where $\mathcal{L}_{CE}^{i}$ is the cross-entropy loss of $i$-th facet, $\lambda$ is a hyper-parameter controlling the weight of contrastive loss, $n$ is the total number of facets.

\section{Experiments}

\subsection{Dataset and Evaluation Metrics}

We conduct experiments on the MITweet \citep{liu-etal-2023-ideology} dataset, which contains 12,594 English tweets and covers 14 highly controversial topics in recent years. Each instance in MITweet is annotated with a relevance label and an ideology label (if the relevance label is ``Related'') for each of the 12 facets in the multifaceted ideology schema. The statistics of MITweet is shown in Table~\ref{tab: dataset}.

We follow the original training/validation/test split and use the same evaluation metrics as \citet{liu-etal-2023-ideology}. First we calculate the Accuracy (Acc) and F1 score for each facet. Then we utilize both \textit{Macro} and \textit{Micro} methods to integrate metrics from all facets to obtain overall results of model performance. Macro-F1 and Macro-Acc are calculated by averaging F1 and Acc across all facets. Micro-F1 and Micro-Acc are the aggregated F1 and Acc scores obtained by concatenating the predictions of all facets. Note that, following existing work, we only report F1-related metrics for \textit{Relevance Recognition} due to the highly imbalanced data distribution in this subtask.

\subsection{Implement Details}

The pre-trained BERTweet-base \citep{nguyen2020bertweet} is used as the concept and text encoder, and the two encoders share weights as this gave better results in preliminary experiments. We train the Relevance Recognition model and the Ideology Analysis model independently. Each model includes the BICo module and is trained end-to-end. We use AdamW \citep{loshchilov2018decoupled} as the optimizer. The learning rate is set to 2e-5. The batch size $B$ is set to 64. The iteration number of BICo is set to 4 for relevance recognition and 2 for ideology analysis. For contrastive loss, we set the temperature parameter $\tau$ to 0.5 for relevance recognition and 0.1 for ideology analysis. The contrastive loss weight $\lambda$ is set to 0.3 for both subtasks. The classification head is a two-layer fully connected network, in which the hidden size is 512. The above parameters are selected based on the validation set. We report the average results of 5 runs with different random seeds.

\subsection{Comparison Models}

We compare our approach with the latest benchmark in the MID task, BERTweetInd \citep{liu-etal-2023-ideology}, which uses BERTweet as the backbone and detects indicator words from training set as the textual descriptions of facets. In addition, we test the zero/few-shot performance of advanced large language models (LLMs) in this task. Specifically, we select two popular LLMs, LLaMA2 \citep{touvron2023llama} and ChatGPT \footnote{\url{https://openai.com/blog/chatgpt}}, which exhibit superior capacities in communicating with humans, including solving a wide range of complex tasks without further training. We use the \texttt{Llama-2-13b-chat} and \texttt{gpt-3.5-turbo-1106} versions. The prompts designed for LLMs can be found in Appendix~\ref{sec: prompt}.

We also provide variants of our proposed approach in the ablation study:

\begin{table}[]
	\centering
	\fontsize{8.5pt}{9.2pt}\selectfont
	\renewcommand\arraystretch{1.25}
	\begin{tabular}{@{}m{0.154\textwidth}<{\raggedright}m{0.056\textwidth}<{\centering}m{0.056\textwidth}<{\centering}m{0.056\textwidth}<{\centering}m{0.056\textwidth}<{\centering}@{}}
		\toprule
		\textbf{Model} & \textbf{Macro-F1}       & \textbf{Micro-F1}       & \textbf{Macro-Acc}      & \textbf{Micro-Acc}      \\ \midrule
		\multicolumn{5}{c}{\textit{Subtask 1: Relevance Recognition}}                                        \\ \midrule
		BERTweetInd              & 57.48          & 70.32          & -              & -              \\ \midrule
		LLaMA2-13B$^\circ$            & 27.45          & 32.28          & -              & -              \\
		ChatGPT$^\circ$              & 33.11          & 40.07          & -              & -              \\
		LLaMA2-13B$^\triangle$            & 29.35          & 38.17          & -              & -              \\
		ChatGPT$^\triangle$              & 38.83          & 44.78          & -              & -              \\ \midrule
		\textbf{Our approach} & $\textbf{59.22}^\dagger$ & $\textbf{72.14}^\dagger$ & -              & -              \\
		w/o CL                & 58.56          & 71.42          & -              & -              \\
		w/o BICo            & 58.14          & 70.41          & -              & -              \\
		w/o CL\&BICo        & 57.85          & 70.42          & -              & -              \\ \midrule
		\multicolumn{5}{c}{\textit{Subtask 2: Ideology Analysis}}                                           \\ \midrule
		BERTweetInd              & 42.68          & 69.28          & 65.88          & 76.38          \\ \midrule
		LLaMA2-13B$^\circ$            & 35.60          & 47.33          & 45.98          & 49.69          \\
		ChatGPT$^\circ$               & 37.11          & 53.41          & 48.57          & 57.95          \\
		LLaMA2-13B$^\triangle$            & 38.51          & 47.22          & 46.13          & 48.90          \\
		ChatGPT$^\triangle$               & 42.64          & 60.54          & 58.44          & 68.25          \\ \midrule
		\textbf{Our approach} & $\textbf{47.32}^\dagger$ & $\textbf{70.90}^\dagger$ & \textbf{66.79} & $\textbf{78.60}^\dagger$ \\
		w/o BICo            & 46.02          & 68.58          & 66.18          & 76.63          \\
		w/o Concept anchors   & 45.08          & 68.38          & 66.04          & 77.30          \\
		w/o CGCL & 44.21          & 67.54          & 65.15          & 76.79          \\ \bottomrule
	\end{tabular}
	\caption{Overall results of different models and ablation study. $\circ$ and $\triangle$ denote 0-shot and 3-shot, respectively. $\dagger$ denotes the significance test over BERTweetInd at p-value<0.05. \textbf{Bold} values are the best results in the corresponding subtask.}
	\label{tab:main results}
\end{table}

\textbullet \ \ \textit{Relevance Recognition}

(1) ``w/o CL'' denotes without contrastive learning.

(2) ``w/o BICo'' denotes without bidirectional iterative concept flow, in which case the facet representations in Concept Attentive Matching are directly from the concept encoder.

(3) ``w/o CL\&BICo'' denotes the combination of the above two cases.

\textbullet \ \ \textit{Ideology Analysis}

(1) ``w/o BICo'' denotes without bidirectional iterative concept flow. In this case, the concept anchors (i.e., ideology representations) are directly from the concept encoder.

(2) ``w/o concept anchors'' denotes performing the contrastive learning without the guidance of concept anchors, i.e., the anchors are text representations themselves, which is the case of Eq.~\eqref{eq11}.

(3) ``w/o CGCL'' denotes discarding the concept-guided contrastive learning.

\begin{table*}[]
	\centering
	\fontsize{8.8pt}{9.2pt}\selectfont
	\renewcommand\arraystretch{1.28}
	\resizebox{\textwidth}{!}{%
		\begin{tabular}{@{}lrrrrrrrrrrrr@{}}
			\toprule
			\multicolumn{1}{c}{Model}  & \multicolumn{1}{c}{PoR} & \multicolumn{1}{c}{SS} & \multicolumn{1}{c}{EO} & \multicolumn{1}{c}{EE} & \multicolumn{1}{c}{EP} & \multicolumn{1}{c}{CSR} & \multicolumn{1}{c}{CV} & \multicolumn{1}{c}{DS} & \multicolumn{1}{c}{MF} & \multicolumn{1}{c}{SD} & \multicolumn{1}{c}{JO} & \multicolumn{1}{c}{PeR} \\ \midrule
			\multicolumn{13}{c}{\textit{Subtask 1: Relevance Recognition}}                                                                                                                                                                                                                                                                       \\ \midrule
			BERTweetInd           & \underline{46.92}                   & \underline{32.71}                  & \underline{71.05}                  & \underline{63.29}                  & \underline{82.26}                  & \textbf{35.04}          & \textbf{19.52}         & \underline{62.73}                  & \underline{85.99}                  & \underline{44.07}                  & \underline{75.55}                  & \underline{70.71}                   \\
			LLaMA2-13B$^\triangle$            & 3.33                    & 9.41                   & 31.30                  & 20.48                  & 47.23                  & 5.19                   & 4.33                   & 30.82                  & 56.42                  & 26.32                  & 56.06                  & 61.34                   \\
			ChatGPT$^\triangle$               & 6.48                    & 10.45                   & 54.27                  & 37.33                  & 53.04                  & 10.27                   & 7.96                   & 47.02                  & 78.87                  & 33.52                  & 63.92                  & 62.79                   \\
			\textbf{Our approach} & $\textbf{52.63}^\dagger$          & $\textbf{33.33}^\dagger$         & $\textbf{71.65}^\dagger$         & $\textbf{67.98}^\dagger$         & \textbf{83.00}         & \underline{31.58}                   & \underline{19.35}                  & $\textbf{66.80}^\dagger$         & $\textbf{86.59}^\dagger$         & $\textbf{50.55}^\dagger$         & $\textbf{76.16}^\dagger$         & $\textbf{70.98}^\dagger$          \\ \midrule
			\multicolumn{13}{c}{\textit{Subtask 2: Ideology Analysis}}                                                                                                                                                                                                                                                                          \\ \midrule
			BERTweetInd           & 24.40                   & 27.59                  & \underline{52.26}         & 41.25                  & 52.43                  & \textbf{49.93}          & 43.37                  & \underline{57.00}                  & \textbf{48.39}         & \underline{43.92}                  & 36.55                  & 35.04                   \\
			LLaMA2-13B$^\triangle$           & \textbf{37.23}                   & \underline{44.47}                  & \textbf{52.28}                  & 36.97                  & 45.56                  & \underline{44.81}                   & 29.60                  & 41.60                  & 29.87                  & 39.90                  & 33.47                  & 26.40                   \\
			ChatGPT$^\triangle$               & 24.44                   & 43.91                  & 50.79                  & \underline{43.73}                  & \underline{54.06}                  & 33.33                   & \textbf{51.59}                  & 52.92                  & 33.03                  & 36.03                  & \textbf{41.95}                  & \textbf{45.87}                   \\
			\textbf{Our approach} & \underline{33.16}$^\dagger$          & $\textbf{45.18}^\dagger$         & 50.25                  & $\textbf{61.70}^\dagger$         & $\textbf{59.65}^\dagger$         & 35.56                   & \underline{49.31}$^\dagger$         & $\textbf{62.35}^\dagger$         & \underline{47.25}                  & $\textbf{45.10}^\dagger$         & \underline{37.10}         & \underline{41.24}$^\dagger$          \\ \bottomrule
		\end{tabular}%
	}
	\caption{F1 scores of different models on each facet. $\dagger$ denotes the significance test over BERTweetInd at p-value<0.05. $\triangle$ denotes 3-shot. \textbf{Bold} and \underline{underlined} values are the best and second-best results in the subtask, respectively. Full names of 12 facets in the first row can be found in Appendix~\ref{sec: schema}.}
	\label{tab: facet results}
\end{table*}

\subsection{Main Results}

We present the overall results of our approach and other models in Table~\ref{tab:main results}. First, we can observe that our concept-enhanced method performs consistently better than other baseline models, including the advanced large language models, indicating the superiority of our approach for the MID task. Second, compared with BERTweetInd, which is also a BERTweet-based model, our approach achieves significant improvements in both subtasks. This suggests that the application of concept semantics in the hierarchical schema helps the model to capture the correlation between text and labels, thus improving the performance. Third, for the LLMs, although ChatGPT performs better than LLaMA2-13B, and the few in-context demonstrations improve the results, there is still a large gap between LLMs and other task-specific models. This indicates that the MID task remains challenging for current LLMs. One possible reason is that, this task requires not only strong text understanding and semantic reasoning abilities, but also the integration of specialized sociological knowledge and background information on relevant topics, which is difficult for general-purpose LLMs.

In more detail, F1 scores of different models on each facet are shown in Table~\ref{tab: facet results}. In the subtask of \textit{Relevance Recognition}, our approach achieves the best results on 10 out of 12 facets, surpassing the second-place by over 4 points on 4 facets (PoR, EE, DS, SD). This again demonstrates the effectiveness of our concept-enhanced framwork in the MID task. However, on the facets of CSR and CV, our approach is inferior to BERTweetInd, especially on CSR. We think this is likely because there are too few related samples in CSR (as shown in Table~\ref{tab: dataset}), and our method uses a separate classification head for each facet, resulting in even more insufficient training for CSR. Although this issue affects the results, it is only an edge case. The two LLMs still perform poorly, especially on PoR, SS and CV. By analyzing the responses generated by LLMs, we find that LLMs are more likely to ignore or generalize the definitions in prompts on these facets. For the \textit{Ideology Analysis} subtask, the baseline models achieve the best or second-best result on some facets. Nevertheless, our approach ranks in the top two on 10 out of 12 facets and shows overall superior performance.

\subsection{Ablation Study}

We conduct ablation studies to inspect the importance of major components in our model and the results are reported in Table~\ref{tab:main results}. It is clear that the removal of either one of the modules causes a drop in performance. The Micro-F1 decreases by 1.73 and 2.32 points on the two subtasks, respectively, when BICo is removed, which validates that it is important to further model the schema hierarchy and concept interactions on top of the concept encoder. BICo iteratively performs concept diffusion and aggregation on the hierarchy tree, and the updated concept representations are enriched by higher-level general semantics and lower-level concrete perspectives, which are helpful for the model to understand the deep meaning of facet and ideology labels.

In \textit{Ideology Analysis}, the removal of concept anchors leads to noticeable performance degradation. This suggests that relying solely on text content to identify ideology is insufficient, and injecting label semantics can guide the model to capture ideology features and distinguish among different ideologies more accurately, so as to improve the performance of MID. Moreover, the results of ``w/o CL'' in \textit{Relevance Recognition} and ``w/o CGCL'' in \textit{Ideology Analysis} verify the effectiveness of contrastive learning strategies in two subtasks.

We also conduct ablation study for the modules of Concept Metapath Diffusion and Concept Hierarchy Aggregation in BICo. The results are presented in Appendix~\ref{sec: ablation}.

\begin{table}[]
	\centering
	\fontsize{8pt}{8.5pt}\selectfont
	\renewcommand\arraystretch{1.25}
	\begin{tabular}{@{}m{0.07\textwidth}<{\centering}m{0.105\textwidth}<{\raggedright}@{\hspace{0.24cm}}m{0.088\textwidth}<{\centering}@{\hspace{0.27cm}}m{0.078\textwidth}<{\centering}@{\hspace{0.27cm}}m{0.068\textwidth}<{\centering}@{}}
		\toprule
		\multirow{2}{*}[-0.12cm]{Test Topics} & \multicolumn{1}{c}{\multirow{2}{*}[-0.12cm]{Model}} & Relevance Recognition & \multicolumn{2}{c}{Ideology Analysis} \\ \cmidrule(l){3-5}
		& \multicolumn{1}{c}{}                       & Micro-F1              & Micro-Acc          & Micro-F1          \\ \midrule
		\multirow{4}{*}{CHR\&GF}     & BERTweetInd                                   & 59.60                 & 70.20              & 52.41             \\
		& LLaMA2-13B                                     & 28.29                 & 56.79              & 44.22             \\
		& ChatGPT                                    & 36.20                 & 69.70              & 51.87             \\
		& \textbf{Our approach}                      & $\textbf{61.00}^\dagger$        & $\textbf{72.43}^\dagger$     & $\textbf{54.76}^\dagger$    \\ \midrule
		\multirow{4}{*}{BLM\&Dm}     & BERTweetInd                                   & 54.69                 & 80.64              & 58.89             \\
		& LLaMA2-13B                                     & 31.90                 & 58.93              & 46.27             \\
		& ChatGPT                                    & 39.54                 & 73.45              & 54.09             \\
		& \textbf{Our approach}                      & $\textbf{62.88}^\dagger$        & $\textbf{83.31}^\dagger$     & $\textbf{61.04}^\dagger$    \\ \bottomrule
	\end{tabular}
	\caption{Cross-topic results of different models. CHR means Capitol Hill Riot, GF means George Floyd, BLM means Black Lives Matter, Dm means Democracy. $\dagger$ denotes the significance test over BERTweetInd at p-value<0.05. \textbf{Bold} values are the best results in the corresponding test topics.}
	\label{tab: cross-topic}
\end{table}

\begin{figure*}[t]
	\centering  
	\subfigure[w/o CGCL]{
		\label{Fig.sub.1}
		\includegraphics[width=0.32\textwidth]{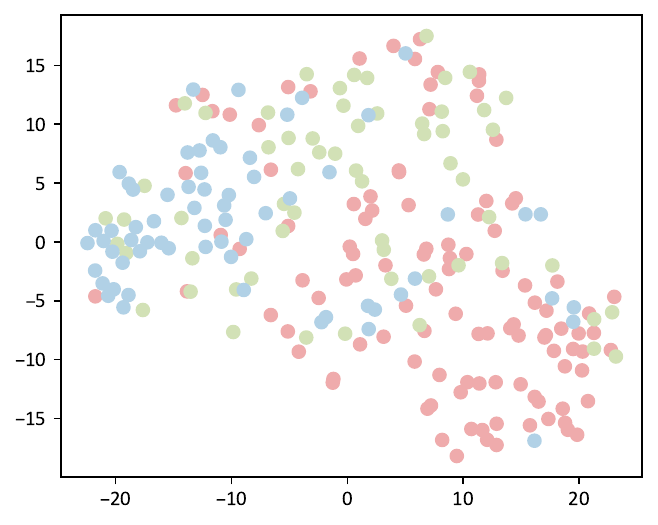}}
	\subfigure[w/ CL]{
		\label{Fig.sub.2}
		\includegraphics[width=0.32\textwidth]{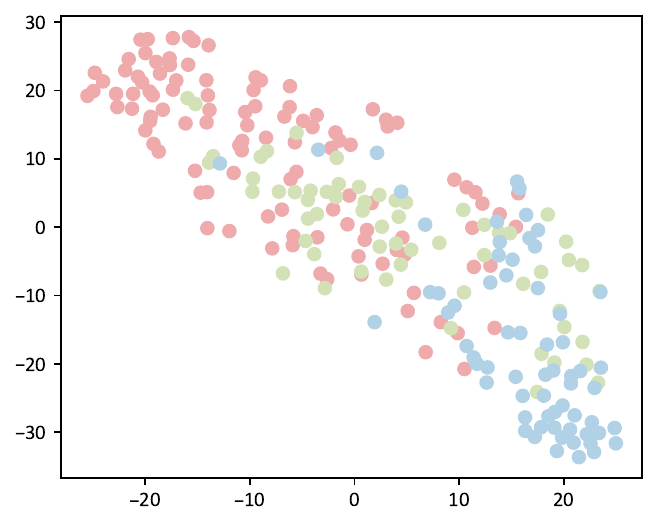}}
	\subfigure[w/ CGCL]{
		\label{Fig.sub.3}
		\includegraphics[width=0.32\textwidth]{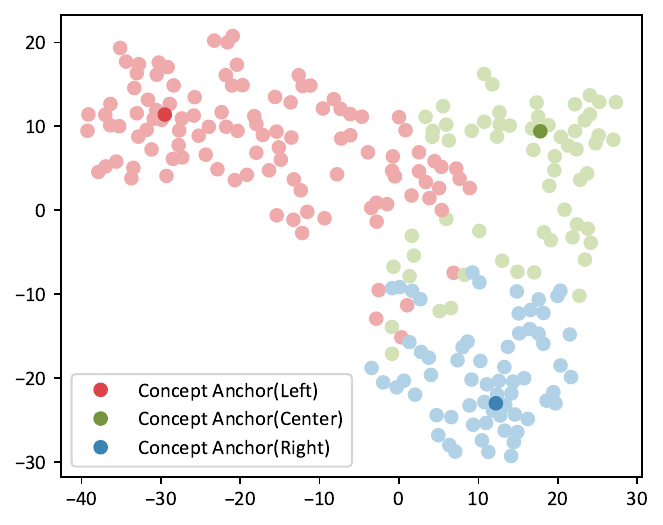}}
	\caption{T-SNE visualization of text representations learned by different model variants in the \textit{Ideology Analysis} subtask. CGCL denotes our Concept-Guided Contrastive Learning. CL denotes the Contrastive Learning without concept anchors. Red, green and blue dots represent Left, Center and Right samples, respectively.}
	\label{fig: visualization}
\end{figure*}

\begin{figure}[t] 
	\centering 
	\includegraphics[width=\columnwidth]{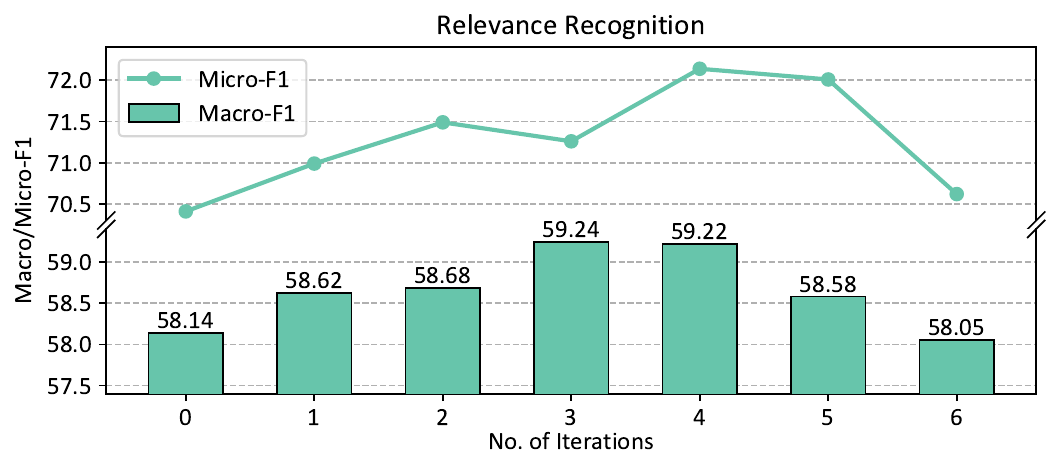} 
	\includegraphics[width=\columnwidth]{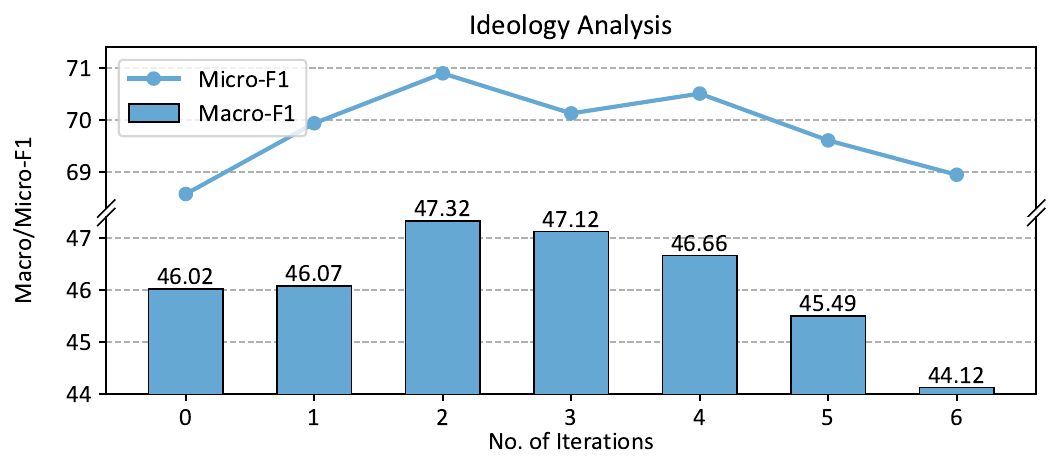} 
	\caption{Results of different numbers of iterations.} 
	\label{fig: iteration} 
\end{figure}

\subsection{Cross-Topic Generalization}
In our approach, label concepts are incorporated to enhance the model and they are enriched by multi-granularity concepts from different levels in the hierarchical schema through BICo. Intuitively, concepts provide a general description of a label. Therefore, our model should have better generalization to new topics with the help of concept semantics. To validate this viewpoint, we test and compare the cross-topic generalization ability of different models.

In the cross-topic scenario, the models are trained on some topics and then tested on the rest topics. To reduce randomness, we conduct experiments on two sets of test topics and the results are shown in Table~\ref{tab: cross-topic}. It can be observed that our approach consistently outperforms other models in both subtasks, which verifies that our approach can better generalize the learning ability to deal with cross-topic scenarios. LLMs lag behind other models by a significant margin. This shows that task-specific models still have advantages even in cross-topic scenarios. However, for the test topics of CHR\&GF, ChatGPT performs closely to task-specific models in the \textit{Ideology Analysis} subtask, indicating that ChatGPT may have practical value in specific cross-topic scenarios.

\subsection{Effect of Number of Iterations}

To analyze the effect of using different numbers of iterations in BICo, we conduct experiments on both subtasks and present the results in Figure~\ref{fig: iteration}. We can observe a clear upward and then downward trend in model performance as the number of iterations increases. The optimal number of iterations for \textit{Relevance Recognition} is 4 and for \textit{Ideology Analysis} is 2. One possible reason for this trend is that, when the number of iterations is too small, the concept diffusion and aggregation are insufficient, and the concept representations do not fully perceive the semantics of different granularities in the hierarchical structure. In contrast, when the number of iterations is too large, there will be redundancy in information transfer, and the semantic features of the concept itself will be lost.

\subsection{Visualization}

To qualitatively examine the role of label semantics (concept anchors) in the concept-guided contrastive learning, we randomly select a facet (Diplomatic Strategy) and show the t-SNE projections of text representations from test set in Figure~\ref{fig: visualization}. As observed, for the case of ``w/o CGCL'', all samples are almost scattered without separations. There is a similar but better distribution for the model trained with CL. While for our CGCL (i.e., the full model), instances are well clustered by labels with only a slight overlap and the concept anchors are approximately cluster centers. This confirms that concept representations learned from BICo guide the model to better distinguish among different ideologies in the embedding space, which is helpful for subsequent classification.

\section{Related Work}

\paragraph{Ideology Detection} This task detects the ideology of texts in a generic facet. Many studies rely on text analysis techniques and try to leverage various textual cues \citep{bhatia-p-2018-topic, baly-etal-2019-multi, baly-etal-2020-detect, chen-etal-2020-analyzing, kabir2022deep, liu-etal-2022-politics, kim-johnson-2022-close, devatine-etal-2023-integrated, hong-etal-2023-disentangling, chen-etal-2023-ideology}. In addition to text content, social networks \citep{li-goldwasser-2019-encoding, stefanov-etal-2020-predicting, xiao-2020-timme, li-goldwasser-2021-using}, external knowledge \citep{kulkarni-etal-2018-multi, zhang-etal-2022-kcd} and multimodal information \citep{dinkov2019predicting, qiu-etal-2022-late} are utilized to identify the ideology of online texts.

\paragraph{Multifaceted Ideology Detection} Considering that some texts may contain descriptions of different issues and reflect the author’s ideology from various aspects, some recent work study ideology detection on multiple facets. \citet{sinno-etal-2022-political} investigate the political ideology of news articles from three facets, social, economic and foreign. \citet{liu-etal-2023-ideology} first propose the MID task and design the first multifaceted ideology schema which defines 5 domains and 12 facets in a hierarchical structure. They also manually annotate a high-quality MITweet dataset and build baselines for MID. We follow \citet{liu-etal-2023-ideology} and introduce label semantics into models through encoding the hierarchical schema.

\section{Conclusion}

In this paper, we have proposed a concept semantics-enhanced framework for the MID task. We have also designed a novel bidirectional iterative concept flow method to capture multi-granularity concept semantics. Moreover, we have explored concept attentive matching and concept-guided contrastive learning strategies to enable the model to extract ideology features with the help of concept semantics. Experiment results have validated the superiority of our approach.

\subsection*{Acknowledgement} This work is supported in part by Major Project of National Social Science Foundation of China: “AI and Precise International Communication” (Grant No. 22\&ZD317) and National Natural Science Foundation of China (Grant No. 62172167). The computation is supported by the HPC Platform of Huazhong University of Science and Technology.
\section*{Limitations}

\begin{itemize}
	\item{Following \citet{liu-etal-2023-ideology}, we divide multifaceted ideology detection into two subtasks in a pipeline manner. However, this modeling approach increases the computational cost in both training and inference stages. In addition, error propagation in this pipeline mode is also a problem that cannot be ignored. We will investigate how to solve this task in an end-to-end manner in future work.}
	
	\item{While we attempt to tune the concepts defined in the schema to better fit our approach, we are constrained by computational resources and time, so we directly adopt the concepts in the schema. Although these concepts are representative, there may be better ones that could lead to better performance.}
\end{itemize}

\section*{Ethical Considerations}

We carry out this work and conduct the experiments in accordance with the general ethics in social science research. The proposed concept-enhanced framework could automatically detect the multifaceted ideology of given texts, which is helpful for policy-makers and social statisticians. However, the algorithm is not perfect and may make incorrect predictions. Therefore, researches should realize the potential harm from the misuse of the ideology detection system, and cannot rely solely on the system to make judgments.

\bibliography{custom}

\appendix

\begin{table*}[]
	\centering
	\small
	\renewcommand\arraystretch{1.2}
	\resizebox{0.85\textwidth}{!}{%
		\begin{tabular}{@{}lm{0.25\textwidth}<{\raggedright}ll@{}}
			\toprule
			\textbf{Domain}            & \textbf{Facet}               & \textbf{Left}         & \textbf{Right}       \\ \midrule
			\multirow{2}{*}{Politics}  & Political Regime (PoR)       & Socialism             & Capitalism           \\
			& State Structure (SS)         & Centralism            & Federalism           \\ \midrule
			\multirow{2}{*}{Economy}   & Economic Orientation (EO)    & Command Economy       & Market Economy       \\
			& Economic Equality (EE)       & Outcome Equality      & Opportunity Equality \\ \midrule
			\multirow{3}{*}{Culture}   & Ethical Pursuit (EP)         & Ethical Liberalism    & Ethical Conservatism \\
			& Church-State Relations (CSR) & Secularism            & Caesaropapism        \\
			& Cultural Value (CV)          & Collectivism          & Individualism        \\ \midrule
			\multirow{2}{*}{Diplomacy} & Diplomatic Strategy (DS)     & Globalism             & Isolationism         \\
			& Military Force (MF)          & Militarism            & Pacifism             \\ \midrule
			\multirow{3}{*}{Society}   & Social Development (SD)      & Revolutionism         & Reformism            \\
			& Justice Orientation (JO)     & Result Justice        & Procedural Justice   \\
			& Personal Right (PeR)         & Social Responsibility & Individual Right     \\ \bottomrule
		\end{tabular}%
	}
	\caption{Multifaceted ideology schema \citep{liu-etal-2023-ideology}.}
	\label{tab: multifaceted ideology schema}
\end{table*}

\section{Multifaceted Ideology Schema}
\label{sec: schema}

We present the multifaceted ideology schema in Table~\ref{tab: multifaceted ideology schema}. Concepts of facets and ideologies defined in the schema can be found in \citet{liu-etal-2023-ideology}. Note that the original schema does not give the concepts of ``Center'', so we define them based on the concepts of ``Left'' and ``Right'', as follows:

\subsection{Domain 1: Politics}

\begin{itemize}[leftmargin=*]
	\item{\textbf{Political Regime (PoR)}
		
	Center: A moderate stance advocating for a mix of public and private ownership, seeking a balanced approach to property control and means of production.}
	\item{\textbf{State Structure (SS)}
		
	Center: A moderate stance advocating for a balanced power structure, combining elements of central authority and power distribution.}
\end{itemize}

\subsection{Domain 2: Economy}

\begin{itemize}[leftmargin=*]
	\item{\textbf{Economic Orientation (EO)}}
	
	Center: A moderate stance advocating for combining government intervention in important economic decisions with the role of individuals, organizations, and market interactions.
	
	\item{\textbf{Economic Equality (EE)}}
	
	Center: A moderate position advocating for an economic system that balances equal treatment and access to resources with considerations for distribution outcomes among different groups.
\end{itemize}

\subsection{Domain 3: Culture}

\begin{itemize}[leftmargin=*]
	\item{\textbf{Ethical Pursuit (EP)}}
	
	Center: The mainstream culture should consider individual freedoms and cultural norms while promoting inclusivity dialogue on controversial issues.
	
	\item{\textbf{Church-State Relations (CSR)}}
	
	Center: A moderate position advocating for a balanced and cooperative relationship between the church and state, respecting both religious autonomy and the principles of secular governance.
	
	\item{\textbf{Cultural Value (CV)}}
	
	Center: A moderate stance that recognizes the importance of both social collectives and individual autonomy in shaping and preserving a diverse and inclusive society.
\end{itemize}

\subsection{Domain 4: Diplomacy}

\begin{itemize}[leftmargin=*]
	\item{\textbf{Diplomatic Strategy (DS)}}
	
	Center: A moderate position that balances international cooperation and national interests, recognizing the value of engagement while cautiously managing political and economic entanglements with other countries.
	
	\item{\textbf{Military Force (MF)}}
	
	Center: A moderate stance that recognizes the need for armed defense and security while prioritizing non-violent resolution for conflicts.
\end{itemize}

\subsection{Domain 5: Society}

\begin{itemize}[leftmargin=*]
	\item{\textbf{Social Development (SD)}}
	
	Center: A moderate position that advocates combining direct action when necessary with a recognition of the value of gradual and sustainable change to achieve social goals.
	
	\item{\textbf{Justice Orientation (JO)}}
	
	Center: A moderate stance that seeks a balance between fair distribution and fair decision-making, considering both the outcomes and procedure of justice.
	
	\item{\textbf{Personal Right (PeR)}}
	
	Center: A moderate position that recognizes the importance of both fulfilling individual responsibilities and protecting individual rights in an equitable manner.
\end{itemize}

\section{Prompts for LLMs}
\label{sec: prompt}

The prompt templates designed for LLMs in two subtasks are as follows. We fill the templates with the facet names and definitions in the multifaceted ideology schema. In few-shot experiments, we provide LLMs with a few in-context demonstrations, which are manually selected for each facet to ensure diversity. We also provide a brief analysis as chain-of-thought for each demonstration. In zero-shot experiments, the demonstrations in the prompts will be removed.

\subsection{Relevance Recognition}

\begin{itemize}[leftmargin=*]
	\item{\textit{System prompt}
		
		You will be provided with a piece of text. Determine if the text is related to "\{facet\}".
		
		\{facet\} is defined as: \{facet\_def\}
		
		First give your analysis briefly and then select your answer from ["Related", "Unrelated"].}
		
		Here are some demonstrations:\\
		\{demonstrations\}
	
	\item{\textit{User prompt}
		
		Text: """\{text\}"""}
\end{itemize}

\subsection{Ideology Analysis}

\begin{itemize}[leftmargin=*]
	\item{\textit{System prompt}
		
		You will be provided with a piece of text. Determine the orientation of the text towards "\{facet\}".
		
		The orientation towards "\{facet\}" can be divided into ["Left", "Right", "Center"]. The definitions are as follows:
		
		-Left: \{left\_def\}\\
		-Right: \{right\_def\}\\
		-Center: \{center\_def\}
		
		First give your analysis briefly and then select your answer from ["Left", "Right", "Center"].}
		
		Here are some demonstrations:\\
		\{demonstrations\}
	
	\item{\textit{User prompt}
		
		Text: """\{text\}"""}
\end{itemize}

\section{Additional Ablation Study}
\label{sec: ablation}

As shown in Table~\ref{tab: additional ablation}, the removal of Concept Metapath Diffusion or Concept Hierarchy Aggregation causes a drop in performance. And removing both of them (w/o BICo) leads to a more significant performance degradation. The concept diffusion from root to leaf enables the high-level general semantics to propagate to lower-level nodes, while the concept aggregation from leaf to root allows the high-level nodes to perceive multifaceted concepts from lower levels. Both contribute to enriching label representations. The results further validate the effectiveness of both modules.

\begin{table}[]
	\centering
	\small
	\renewcommand\arraystretch{1.25}
	\begin{tabular}{@{}m{0.146\textwidth}<{\raggedright}m{0.058\textwidth}<{\centering}m{0.058\textwidth}<{\centering}m{0.058\textwidth}<{\centering}m{0.058\textwidth}<{\centering}@{}}
		\toprule
		\textbf{Model}        & \textbf{Macro-F1} & \textbf{Micro-F1} & \textbf{Macro-Acc} & \textbf{Micro-Acc} \\ \midrule
		\multicolumn{5}{c}{\textit{Subtask 1: Relevance Recognition}}                                           \\ \midrule
		\textbf{Our Approach} & \textbf{59.22}    & \textbf{72.14}    & \textbf{-}         & \textbf{-}         \\ \midrule
		w/o CMD               & 57.92             & 70.81             & -                  & -                  \\
		w/o CHA               & 58.73             & 71.03             & -                  & -                  \\
		w/o BICo              & 58.14             & 70.41             & -                  & -                  \\ \midrule
		\multicolumn{5}{c}{\textit{Subtask 2: Ideology Analysis}}                                               \\ \midrule
		\textbf{Our Approach} & \textbf{47.32}    & \textbf{70.90}    & \textbf{66.79}     & \textbf{78.60}     \\ \midrule
		w/o CMD               & 46.38             & 68.93             & 65.90              & 77.71              \\
		w/o CHA               & 46.13             & 69.48             & 66.75              & 77.85              \\
		w/o BICo              & 46.02             & 68.58             & 66.18              & 76.63              \\ \bottomrule
	\end{tabular}
	\caption{Results of ablation study for the modules of Concept Metapath Diffusion (CMD) and Concept Hierarchy Aggregation (CHA) in BICo. Note that ``w/o BICo'' is equivalent to ``w/o CMD\&CHA''}
	\label{tab: additional ablation}
\end{table}

\begin{table}[]
	\centering
	\renewcommand\arraystretch{1.06}
	\resizebox{0.97\columnwidth}{!}{%
		\begin{tabular}{@{}ll|r|rrr@{}}
			\toprule
			\multicolumn{1}{l}{\multirow{2}{*}[-0.12cm]{\textbf{Domain}}} & \multicolumn{1}{c|}{\multirow{2}{*}[-0.12cm]{\textbf{Facet}}} & \multicolumn{1}{c|}{\textbf{Relevance}} & \multicolumn{3}{c}{\textbf{Ideology}}                  \\ \cmidrule(l){3-6}
			\multicolumn{1}{c}{}                                 & \multicolumn{1}{c|}{}                                & \textbf{\#Related}                      & \textbf{\#Left} & \textbf{\#Center} & \textbf{\#Right} \\ \midrule
			\multirow{2}{*}{Politcs}                             & PoR                                                  & 112                                     & 39              & 14                & 59               \\
			& SS                                                   & 291                                     & 67              & 88                & 136              \\ \midrule
			\multirow{2}{*}{Economy}                             & EO                                                   & 759                                     & 294             & 297               & 168              \\
			& EE                                                   & 672                                     & 520             & 119               & 33               \\ \midrule
			\multirow{3}{*}{Culture}                             & EP                                                   & 2935                                    & 1976            & 465               & 494              \\
			& CSR                                                  & 68                                      & 33              & 17                & 18               \\
			& CV                                                   & 154                                     & 95              & 11                & 48               \\ \midrule
			\multirow{2}{*}{Diplomacy}                           & DS                                                   & 1572                                    & 711             & 421               & 440              \\
			& MF                                                   & 1837                                    & 132             & 575               & 1130             \\ \midrule
			\multirow{3}{*}{Society}                             & SD                                                   & 1737                                    & 1236            & 287               & 214              \\
			& JO                                                   & 3452                                    & 3058            & 281               & 113              \\
			& PeR                                                  & 3516                                    & 171             & 241               & 3104             \\ \bottomrule
		\end{tabular}%
	}
	\caption{Statistics of the MITweet dataset.}
	\label{tab: dataset}
\end{table}

\end{document}